\documentstyle[a4wide,newlfont]{article}
\title{Active Constraints for a Direct Interpretation of HPSG}
\author{Philippe Blache and Jean-Louis Paquelin\\
2LC - CNRS \\ 1361 route des Lucioles \\ F-06560  Sophia Antipolis \\ {\tt
\{pb,jlp\}@llaor.unice.fr} }
\date{}

\begin{document}
\maketitle

\section{Introduction}

Implementing a linguistic theory raises (at least implicitly) the question of
the interpretation level. \cite{Evans87} distinguishes indirect, weak direct
and strong direct interpretations: the former uses an intermediate mapping
between the original theory and the grammars whereas the latter treats the
grammars as directly characterising the language. Practically, the better the
adequacy between linguistic and computational models, the higher the directness
level\footnote{In the same perspective, \cite{Fong91} completes {\em
directeness} with the notion of {\em faithfulness}.}. An indirect
interpretation compiles the original formalism into another one (e.g. an HPSG
grammar into a simple phrase structure one) in order to apply traditional
parsing techniques.
A strong direct interpretation implements parsing mechanisms as decribed in the
theory.

We argue in this paper that high-level languages can provide a good adequacy
between the theory and its implementation. We approach in particular the
question  of constraints implementation and show how constraint logic
programming (and more particularly multi-paradigm languages such as {\sf\small
LIFE}) constitutes an efficient implementation framework. This paper describes
how such a direct approach preserves the fundamental properties of the
linguistic theory.

\section{Indirect vs. Direct Interpretations}

\begin{center}
\begin{figure}

\begin{tabular}{|c|c|}
\hline \\
\begin{minipage}[b]{5cm}
\begin{picture}(0,0)%
\includegraphics{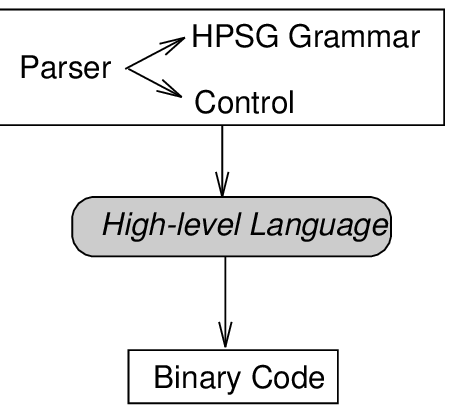}%
\end{picture}%
\setlength{\unitlength}{0.012500in}%
\begin{picture}(143,126)(359,564)
\end{picture}
\end{minipage}
&
\begin{minipage}[b]{5cm}
\begin{picture}(0,0)%
\includegraphics{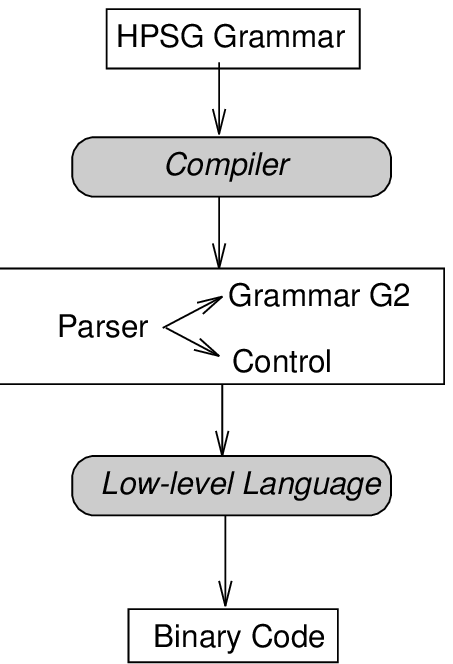}%
\end{picture}%
\setlength{\unitlength}{0.012500in}%
\begin{picture}(143,209)(359,564)
\end{picture}
\end{minipage} \\ \\
{\sf Direct Interpretation} & {\sf Indirect Interpretation} \\
\hline
\end{tabular}
\caption{Systems Architecture}
\label{archi}
\end{figure}
\end{center}

Most of the implementations rely on indirect or weak direct interpretations and
generally compile the original formalism into Prolog clauses (see for example
\cite{Carpenter93}, \cite{Gotz95} or \cite{Popowich91}). We can distinguish two
different approaches according to the level of the implementation. One method
consists of implementing the parser using a high-level language and relying on
mechanisms as close as possible to the theory. In this case, the language is
used both for the knowledge representation (i.e. coding the grammar) and the
implementation of parsing mechanisms. The second approach proposes specific
languages used for representing grammars and generates parsers using a
low-level language\footnote{In the case of ALE, the parser is generated in
Prolog but uses low-level instruction. In fact, as proposed by the authors (see
\cite{Carpenter94}), the host language should be C.}. The figure (\ref{archi})
presents these approaches.

We think that there is a deep difference between them from several point of
view   concerning in particular faithfulness, generality and control. Our
argumentation relies on the observation of the parsers architecture and more
particularly on the specification of the mechanisms required by HPSG. We
distinguish the fundamental characteristics of a theory from the corresponding
operational concepts. As for HPSG, the basic notions are universal principles,
structure sharing, sort hierarchy, well-typedness, etc. Their implementation
requires specific mechanisms such as in particular unification, constraint
propagation, constraint satisfaction, inheritance, underspecification, delayed
evaluation, etc. The adequacy between these two levels seems to be very
important both for linguistics and computational reasons.

\subsection{Faithfulness}

Faithfulness, and more precisely the adequacy between the theoretical model and
its implementation, can be considered as a non formal criterion and evaluated
from a formal and an operational point of view. From a formal point of view, a
faithful approach applies directly the parsing algorithm to the original
grammar whereas an indirect interpretation generates a grammar relying on
another formalism. This difference is not purely esthetic: a direct approach
implements {\em and} validates the model.

As for the operational point of view, we think that faithfulness preserves in
the implementation the properties of the theory itself. Let us precise this
aspect for HPSG by focusing on two important characteristics: {\em generality}
and {\em integration}.

\begin{itemize}
\item {\em Generality} concerns the ability of representing universal
phenomenons. This property is in fact closely related to the reusability which
concerns both the linguistic level (reusing different grammars) and the
computational one (reusing the same parser for different grammars).

A faithful implementation preserves the generality level of the mechanisms (and
their reusability) in the sense that if the theory describes a general
high-level property (e.g. universal principles), then the corresponding
mechanism is at the same level (e.g. active constraint on types). This entails
a distinction between the architecture of the feature structures and the
constraints that they must satisfy: on the one hand, a feature structure must
be totally well-typed (architectural property) and correspond to an ID-schema,
on the other hand, it must satisfy the universal principles. A faithful
implementation using active constraints (in the constraint programming sense)
allows such a distinction whereas a Prolog implementation cannot separate these
levels: in this case, principle verification is evaluated after the
instantiation of the concerned feature structure (for this reason, as in ALE,
universal principles often belong the description of the ID-schematas).

\item The second property, {\em integration}, concerns the ability of
representing in an homogeneous way different source of linguistic knowledge:
prosody, phonology, morphology, syntax, semantics, etc. This property, as for
generality, relies on the distinction between structures and constraints:
integrated approaches must represent various kind of informations within a same
structure. The relations between these informations are described using
constraints. But there is another important aspect concerning the dynamicity of
these relations. Indeed, an integrated approach must describe mutual
dependencies between the different levels of information. These relations have
consequences on the structure itself (in particular via structure sharing), but
also on the processes constructing the corresponding structure. For example,
syntactic informations can have consequences at the phoneme recognition level.
This characteristic entails an on-line process and a direct manipulation of the
original structures.

\end{itemize}
\subsection{Control}

The control problem constitutes another divergent criterion between direct and
indirect approaches. Several aspects can be underlined.

The first point is theoretical and concerns the system architecture. The figure
(\ref{archi}) indicates that the grammar developper in the case of an indirect
system encodes the grammar into a specific formalism. The compiler then
generates the parser itself. It is considered as a black box and the semantic
is not accessible to the grammar developper who has no control on the parser
itself. In the case of direct approaches, the developper knows the semantic of
the language and has a direct control on the parser.

The second aspect concerns more precisely the implementation level. The current
state of parsing technologies shows that we need a clear distinction between
the linguistic knowledge (including architectural aspects) and the
implementation mechanisms. Practically, as shown in the figure (\ref{archi}),
this separation is present in the parsers. The problem comes from the fact that
in the case of an indirect approach, it is not always possible to apply such a
distinction. This is the case for the application of the principles. Their
possible presence  in the ID-schematas is only justified by the fact that
Prolog cannot represent directly constraints on types and must verify such
properties on instantiated object. At the opposite, a language providing active
constraints on types allows the declaration of such constraints a priori, in a
global and persistent way.

We can generalize this last remark to the adequacy between the mechanisms
required by HPSG and those actually implemented. If the host language of the
system doesn't provide adequate mechanisms, they are simulated (e.g.
inheritance becomes an inference process).

Finally, indirect approaches need an interpretation of both formalisms and
mechanisms. We think that before the efficiency problem (which is the main
argument for indirect approaches), the actual problem for the implementation
concerns the preservation of the theory's generality (of great importance in
particular concerning reusability and maintenance).

\section{A Constraint Logic Programming Solution}

As described in the previous section, coding a grammar concerns  knowledge
representation but also interpretation of the implicit mechanisms of the
theory. The question is now: is it possible to directly code a HPSG grammar
into a high-level language or must we use a specific language ? We describe in
this section a solution proposed within the constraint logic programming
paradigm.

\subsection{Active constraints}

HPSG generally considers constraints as descriptions that a well formed object
must satisfy (see \cite{Carpenter92}).
In this definition, the notion of constraint is very precise and restrictive in
comparison with the traditional sense in linguistics. But it has a direct
interpretation within the constraint programming paradigm with active
constraint. We present this notion in this section and compare it with
traditional approaches.

The classical evaluation method in logic programming relies on
generate-and-test: it generates variable values before verifying their
properties. Obviously, a value can be controled by unification with the head of
the predicate, but it is impossible to evaluate properties ({\em i}) before the
unification itself and ({\em ii}) if the object represented by the variable is
only partially known.

Active constraints can implement some of these properties and reduce the search
space by applying them a priori: substitutions are allowed only if the
constraint system remains coherent.
In a constraint logic programming paradigm, the constraint satisfaction
mechanism replaces unification: each resolution step verifies the
satisfiability of the constraint system and simplifies it.
Binding a variable adds new constraints to this system.
In other words, the classical method in Prolog uses a single kind of
constraints (unification) whereas a constraint-based approach allows the
definition of complex ones with a global scope.

Concerning HPSG, a direct interpretation consists of implementing principles
with active constraints.
This approach allows a clear distinction between the basic parsing mechanisms
and the control level.
The parse level consists of determining the possible relations (basically the
valency) whereas constraints verify the well formedness of the structure.
Insofar as constraints are active, such a verification has two main
characteristics: it is an on-line process and it doesn't need any extra
resolution step.
Indeed, a classical Prolog implementation explicitly verifies the
well-formedness using a set of predicates whereas a constraint approach
verifies the satisfiability of the constraint system and unifies two terms in a
single resolution step.
It is clear that the evaluation of the constraint system satisfiability has a
cost, but lower than a classical method because ({\em i}) the system can be
simplified (whereas a classical resolution requires the evaluation of all the
properties) and  ({\em ii}) the search space is reduced a priori (this improves
the control level).

\subsection{Implementation in {\sf LIFE}}

The basic mechanisms required for a direct HPSG interpretation are in
particular unification, constraint satisfaction and inheritance. As for
knowledge representation, the basic objects are the typed feature structures.
The language {\sf\small LIFE} (cf. \cite{Ait-Kaci91}) implements all these
requirements. It is a multi-paradigm language (functional, logic, constraint,
object oriented paradigms) which uses the $\psi$-terms as basic objects.
{\sf\small LIFE} offers  built-in inheritance together with constraint solving
mechanisms: these characteristics allow ({\em i}) to constrain the terms and
({\em ii}) to control  propagation.

The following points are just a sketch illustrating the relevance of this
framework for a direct interpretation.

\subsubsection{Principles}
In this language, constraints are expressed on the sorts: descriptions
corresponding to principles are implemented directly in this way. We can remark
that the formulation of these principles are very similar in all the system
implementing HPSG. The difference here doesn't concern the representation but
the evaluation. We take here the case of two basic principles implemented as
active constraints on the type {\em phrase}. Each term of this type must
satisfy these constraints even if it is not instantiated.

\begin{itemize}
\item {\em HFP: }

{\tt\small
\begin{tabular}{ll}
:: P: phrase $\mid$ & (P.synsem.loc.cat.head = X, \\
	&  P.dtrs.head-dtr.loc.cat.head = X).
\end{tabular}}

Such a constraint stipulates that every term of sort {\em phrase} must have the
concerned head values refering to the same term (tagged by {\tt X}).

\item {\em Valency Principle: }

{\tt\small
\begin{tabular}{ll}
:: P: syntagme $\mid$ ( & P.synsem.loc.cat.subj = X, \\
	& P.dtr.subj-dtr = Y, \\
	& P.dtr.head-dtr.synsem.loc.cat.subj = append(X,Y), \\
	& P.synsem.loc.cat.comps = U, \\
	& P.dtr.comp-dtrs = V, \\
	& P.dtr.head-dtr.synsem.loc.cat.comps = append(U,V)).
\end{tabular}
}

This principle has also a classical formulation, very close to that proposed in
other approaches which is not surprising. What is interesting here is the use
of the function {\tt append} which residuates if its arguments are
unsufficiently known. This constraint can therefore be applied a priori and the
instantiation of one of the concerned features fires the evaluation of the
function and install the constraint.

\end{itemize}

In a classical Prolog implementation, these principles are verified after the
construction of each phrasal sign. In {\sf\small LIFE}, these constraints are
(automatically) satisfied at each moment by these terms.
The satisfiability is not evaluated after the complete instantiation of the
term  but checked at each step since its creation: incoherences are detected
sooner than for classical generate-and-test approaches.

\subsubsection{Inheritance}
Inheritance relations allows the specification (and the propagation) of several
properties.
A well-formed sign must satisfy principles together with these properties.

Inheritance in {\sf\small LIFE} can be seen as a constraint in the sense that
it is integrated to the unification algorithm.
This improves a classical Prolog approach because, as shown in
\cite{Ait-Kaci86}, inheritance is processed by unification steps instead of
resolution ones.

Practically, we implement directly the sort hierarchy using sort inheritance
definitions as described in the theory.

\vspace{2mm}
{\tt\small
\begin{tabular}{ll}
lex <$\mid$ sign. & phrase <$\mid$ sign. \\
noun <$\mid$ subst.& subst <$\mid$ head. \\
...
\end{tabular}}
\vspace{2mm}

Each sort being possibly constrained with a particular property, this mechanism
can implement, as in the theory, complex description relying on multiple
inheritance.

\subsubsection{Sort Resolution and Appropriateness}

HPSG defines both sort hierarchies and features appropriated to each sort.
{\sf\small LIFE} allows to closely follow HPSG's definitions by ({\em i})
defining sort
inheritance hierarchy and ({\em ii}) constraining features associated to them.
For example, we declare the {\em substantive} sort which subsorts are {\em
noun, verb, adjective, preposition and relativizer}. Then we pose the
constraint about the sort {\em category}, which {\sc head} feature must be of
sort lower than {\em substantive}.

\vspace{2mm}
{\tt\small
substantive := {noun;verb;adj;prep;reltvzr}.

:: C:category(head => substantive) | C.head :< substantive.
}
\vspace{2mm}

However, LIFE permits to dynamically enlarge feature structures unlike HPSG
where feature structures are canonical. To constrain LIFE to have at most the
feature appropriated to a structure, we pose some additionnal constraints.

\vspace{2mm}
{\tt\small
:: C:category | lmember(features(C), [head,valence,marking]).
}
\vspace{2mm}

Then, the  sort {\em category} is constrained to have at most the {\sc head,
valence} and {\sc marking} features as defined in the theory. The
\texttt{lmember} predicate parses the authorized features list and succeed if
all features of \texttt{C} belong to it.

\section{Conclusion}

Our approach shows that a strong direct interpretation can be efficient in
several respects.
First, the implementation framework is an actual programming language which
avoid the development of translation tools.
Second, a direct interpretation allows a good maintenance and reusability of
the systems in particular because the generality of the theoretical framework
is preserved.
Finally the constraint programming paradigm offers very efficient properties
useful particularly for knowledge representation and control.
To summarize, the implementation of linguistic constraint using active
constraint is concise, faithful and efficient.


\begin{thebibliography}{12}
\small

\bibitem [A\"{\i}t-Kaci86] {Ait-Kaci86} A\"{\i}t-Kaci H. \& R. Nasr (1986),
``Login: A Logic Programming Language with Built-in Inheritance", in {\em
Journal of Logic Programing}, 1986:3.

\bibitem[A\"{\i}t-Kaci91] {Ait-Kaci91} A\"{\i}t-Kaci H. \& A. Podelski (1991),
Towards a Meaning of {\sf LIFE}, in {\em Proceedings of the 3rd International
Symposium on Programming Language Implementation and Logic Programming},
Springer-Verlag.

\bibitem [Blache95] {Blache95} Blache P. \& N. Hathout (1995) ``Constraint
Logic Programming for NLP", in proceedings of the 5th International Workshop on
{\em Natural Language Understanding and Logic Programming}.

\bibitem [Carpenter92] {Carpenter92} Carpenter B. (1992) {\em The
Logic of Typed Feature Structures}, Cambridge University Press.

\bibitem [Carpenter93] {Carpenter93} Carpenter B. (1993), {\em ALE - The
Attribute Logic Engine. User's Guide}, CMU-LCL Report.

\bibitem [Carpenter94] {Carpenter94} Carpenter B. \& G.  Penn (1994)
``Compiling Typed Attribute-Value Logic Grammars", {\em Technical
Report}, Carnegie Mellon University.


\bibitem [Evans87] {Evans87} Evans R. (1987), {\em Theoretical and
Computational Interpretations of GPSG}, PhD Thesis, University of Sussex.

\bibitem [Fong91] {Fong91} Fong S. (1991), {\em Computational Properties of
Principle-Based Grammatical Theories}, PhD Thesis, MIT.

\bibitem [G\"otz95] {Gotz95} G\"otz T. \& D. Meurers (1995), ``Compiling HPSG
Type Constraints into Definite Clause Programs", in Proceedings of {\em
ACL'95}.

\bibitem [Minnen95] {Minnen95} Minnen G., D. Gerdemann \& T. G\"otz (1995)
``Off-line Optimization for Earley-style HPSG Processing", in proceedings of
{\em EACL'95}.

\bibitem [Pollard \& Sag94] {Pollard94} Pollard C. \& I.  Sag (1994), {\em
Head-driven Phrase Structure Grammars}, CSLI Lecture Notes, Chicago
University Press.

\bibitem [Popowich91] {Popowich91} Popowich F. \& C.  Vogel (1991)
``Logic-Based Implementation of HPSG", in {\em Natural Language
Understanding and Logic Programming III}, C.  Brown \& K.  Koch eds.,
North Holland.

\end{thebibliography}
\end{document}